\documentclass[sigconf,nonacm]{acmart}
\renewcommand\footnotetextcopyrightpermission[1]{}
\usepackage{threeparttable}

\AtBeginDocument{%
  \providecommand\BibTeX{{%
    \normalfont B\kern-0.5em{\scshape i\kern-0.25em b}\kern-0.8em\TeX}}}

\begin{document}

\title{Deep Transfer Learning for Infectious Disease Case Detection Using Electronic Medical Records}


\def\correspondingauthor{\footnote{Corresponding author: yey5@pitt.edu}}

\author{Ye Ye}
\affiliation{%
  \institution{University of Pittsburgh}
  \streetaddress{5607 Baum Blvd, 4th floor}
  \city{Pittsburgh}
  \state{PA}
  \country{USA}}
  \postcode{15206}
  \email{yey5@pitt.edu}

\author{Andrew Gu}
\affiliation{%
  \institution{Carnegie Mellon University}
  \streetaddress{5000 Forbes Ave}
  \city{Pittsburgh}
  \country{PA}}
%\email{andrewg2@andrew.cmu.edu}

%\author{Jonanthon Byrd}
%\affiliation{%
%  \institution{Carnegie Mellon University}
%  \streetaddress{5000 Forbes Ave}
%  \city{Pittsburgh}
%  \country{PA}}
%\email{jonathob@andrew.cmu.edu}


%\renewcommand{\shortauthors}{Ye, Gu, and Byrd.}

\begin{abstract}
   During an infectious disease pandemic, it is critical to share electronic medical records or models (learned from these records) across regions. Applying one region's data/model to another region often have distribution shift issues that violate the assumptions of traditional machine learning techniques. Transfer learning can be a solution. To explore the potential of deep transfer learning algorithms, we applied two data-based algorithms (domain adversarial neural networks and maximum classifier discrepancy) and model-based transfer learning algorithms to infectious disease detection tasks. We further studied well-defined synthetic scenarios where the data distribution differences between two regions are known. Our experiments show that, in the context of infectious disease classification, transfer learning may be useful when (1) the source and target are similar and the target training data is insufficient and (2) the target training data does not have labels. Model-based transfer learning works well in the first situation, in which case the performance closely matched that of the data-based transfer learning models. Still, further investigation of the domain shift in real world research data to account for the drop in performance is needed.
\end{abstract}




%% \begin{teaserfigure}
%%   \includegraphics[width=\textwidth]{sampleteaser}
%%   \caption{Seattle Mariners at Spring Training, 2010.}
%%   \Description{Enjoying the baseball game from the third-base
%%   seats. Ichiro Suzuki preparing to bat.}
%%   \label{fig:teaser} 

%% \end{teaserfigure}

\maketitle
\thispagestyle{empty}

\section{Introduction}
%-Infectious disease case detection using EMR
%-Challenges of medical research using EMR from multiple healthcare systems: NLP, distribution shifting, annotation (test capability)
%-Transfer learning: 
The ability to predict, forecast, and control disease outbreaks highly depends on the ability of disease surveillance. Compared to traditional laboratory reporting and physician reporting systems, automated case detection based on electronic medical records has much less time delay and can cover a large population. However, development of an automated case detection system may require a large training dataset for modeling, which not all regions have available. One solution is that, if region A were affected by an outbreak first, it could share its developed case detection system with region B before region B becomes significantly affected. 

Since both regions cover different populations that are served by healthcare institutions with different electronic medical record systems, their features and distributions for case detection could be different. These differences may not be handled well by traditional machine learning technologies, because they usually work under the assumption that the data for model development and the data for model deployment later have the same underlying distribution, an assumption that cannot hold when the training and test of the model are done at different regions/health care systems. Not only could the test data have a different set of features from the training data, but the correlation between predictive features and class variable could be different between training data and test data. Therefore, when applying a model developed with training data to the test data, the differences between them could lead to a dramatic performance drop.

Transfer learning can be a potential solution. Transfer learning ~\cite{pan2009survey} offers an effective way of reusing data/model (for one task) for another task, by considering both the similarities and the differences between the two tasks. Having been successfully used for many tasks, transfer learning may be able to facilitate data/model sharing to enhance infectious disease case detection. Once region B has a few cases, transfer learning techniques could be used to adapt the case detection system from region A using the data pattern appearing in region B, thereby increasing the case detection capability for region B. 

This study explored the potential of deep transfer learning techniques for infectious disease case detection tasks, which aim to identify cases of infectious respiratory diseases (e.g., influenza) from emergency department visits based on their different clinical manifestations. The patients' clinical signs or symptoms have been extracted from electronic medical records using natural language processing, the use of which further increases the difference between data distributions that need to be handled by transfer learning. We compared different data-based and model-based deep transfer learning algorithms in both real world scenarios and simulated scenarios. The simulation experiments allow us to study how transfer learning generated models perform with respect to the known difference between the source and target distributions.

\section{Background}
Transfer learning is defined as follows : "Given a source domain $D_S$ and learning task $T_S$, a target domain $D_T$, and learning task $T_T$, transfer learning aims to help improve the learning of the target predictive function (target model) $f_{T}(.)$ in $D_T$ using the knowledge in $D_S$ and $T_S$, where $D_{S} \neq D_{T}$, or $T_S \neq T_T$.~\cite{pan2009survey}" In this definition, a domain is denoted by $D = \{X, P(X)\}$, where $X$ is the feature space, and $P(X)$ is the marginal probability distribution of features. A task is denoted by $T = \{Y, f(.)\}$, where $Y$ is the label space of the class variable, and $f(.)$ is an objective predictive function to be learned. Transfer learning has been studied in classic machine learning models, such as logistic regression~\cite{liao2005logistic}\cite{bickel2007discriminative}, SVM~\cite{jebara2004multi}\cite{evgeniou2004regularized}\cite{wu2004improving}, and Bayesian networks~\cite{schwaighofer2004learning}\cite{dai2007transferring}.

%Existing transfer learning techniques can be divided into two branches: data transfer and model transfer. Data transfer includes four main categories: instance weighting (Huang et al. 2006, Jiang 2008, Sugiyama et al. 2008), feature representation (Aue et al. 2005, Arnold et al. 2007, Jiang et al. 2007, Satpal et al. 2007, Ciaramita et al. 2010, Pan et al. 2010, Pan et al. 2011, Wiens et al. 2014, Ogoe et al. 2015), self-labeling (Dai et al. 2007, Tan et al. 2009), and the hyper-parameter strategy (Roy et al. 2007, Finkel et al. 2009).

%add a lot of reference in following two sentences. 
Recent explorations of transfer learning mostly focus on deep neural networks~\cite{deep_transfer_learning_survey}, namely deep transfer learning. Deep transfer learning has been successfully applied in many biomedical image classification tasks~\cite{ribeiro2016exploring}\cite{kermany2018identifying}
\cite{kim2018deep}\cite{mazo2018transfer} and for biomedical name  entity recognition~\cite{sachan2018effective}. Deep transfer learning can be accomplished using either source data or source models.

%~\cite{} introduces four deep transfer learning strategies: instances-based (assigning source instances weights), mapping-based (projecting features from two domains into one common space with better similarity), adversarial-based (applying adversarial approach to create domain invariant features between source and target domains), and network-based (reusing partial network from the source domain). These strategies are not mutually exclusive. Instead, recent deep transfer learning algorithms often use more than one strategies. 
%When considering application scenarios, we may classify deep transfer learning algorithms into two categories: transfer learning that use source data directly (data-based) and that only use source model  (model based):

\textbf{Data-based deep transfer learning} leverages labeled source data along with labeled/unlabeled target data to learn a classifier with strong predictive ability on the target data. There exist several different approaches to using the source data. Mapping-based approaches learn a mapping of the source and target data to a new feature space in which the two are similar, as measured by some distance metric, and then train a classifier using this shared space. An example algorithm that learns the new mapping is Transfer Component Analysis (TCA)~\cite{pan2010domain}. Instance weighting approaches, like TrAdaBoost, find an appropriate weighting for some subset of the source data to be used to augment the target data~\cite{tradaboost}. This is particularly helpful when the domain data size is small. Recently, researchers have also found success in adversarial approaches, where a model is trained to be indiscriminate between the source and target data, while being discriminate for the original predictive task~\cite{deep_transfer_learning_survey}. To achieve this, the training objective combines the original classification loss with an additional loss corresponding to a domain classifier's ability to discriminate between source and target examples. Several state-of-the-art methods combine elements from these different approaches, for example by learning the shared source-target mapping through the inclusion of an adversarial domain classifier~\cite{dann}\cite{mcd}\cite{dada}.
%the available approaches to transfer-learning from source data include instance weighting,7-9 feature representation,10-18 self-labeling,19,20 hyper-parameter strategies,21,22 and deep learning.23-26  [Mention the review paper. Mention the algorithms that people developed.]

\textbf{Model-based deep transfer learning} reuses a partial network from a trained source model and tunes the remaining layers for the target domain, thereby making use of the shareable knowledge from the source~\cite{yosinski2014transferable}. For example, a recent study used a CNN trained on the ImageNet dataset (source model) plus 108,312 optical coherence tomography images (target data) to diagnose the most common blinding retinal diseases (target task)~\cite{kermany2018identifying}. Its results showed that deep transfer learning significantly increased the prediction accuracy and shortened the training time to perform the target task as compared to deep learning using the same target data without using a source model.

\section{Methods}
% By the final report, we expect you to have implemented your own ideas beyond the baseline. Additionally, you should describe what work you have completed towards creating a method which beats the baseline. In addition to successful approaches, you should briefly detail approaches which you tried and found to not work well. What methods have you completed? What is your motivation behind these techniques (you are highly encouraged to come up with an original idea of your own or interesting applications rather than simply implementing or applying existing ML algorithms)?

We compare a data-based deep transfer learning strategy with a model-based deep transfer learning strategy, because the latter may be more feasible for biomedical knowledge sharing across institutional boundaries. In particular, we conduct experiments in both real-world tasks and synthetic scenarios to gain insight into how the difference in the source and target distributions and the availability of target data influence transfer learning performance. 

%In the midway report, we briefly reviewed the two types of transfer learning strategies, identified two data-based algorithms, designed a synthetic data simulation approach, and retrieved real world research data. We then conducted simple experiments to show that neural network models can be used for infectious disease classification tasks. In this report, we explore experiment results in both synthetic scenarios and real-world tasks to gain insight into how the difference in the source and target distributions and the availability of target data influence transfer learning performance. 

\subsection{Data-Based Deep Transfer Learning}
We consider two data-based transfer learning algorithms: domain adversarial neural networks (DANN)~\cite{dann} and maximum classifier discrepancy (MCD)~\cite{mcd}. We choose these two to explore because DANN represents the simplest approach to adversarial transfer learning and MCD recently achieved improvements on a few image domain adaptation tasks~\cite{mcd}. These two algorithms can conduct unsupervised transfer learning scenarios where the class labels of the target data are not available.

\subsubsection{Domain Adversarial Neural Networks}
The DANN approach (Figure \ref{fig:dann-workflow}) uses a single feedforward network, consisting of a feature extractor $G_f$ and label predictor $G_y$ (for the source task if no target label is available), in tandem with a domain classifier $G_d$. The loss function trades off between performing well on the source label prediction task and fooling the domain classifier, as controlled by a coefficient $\lambda$. In that way, the network learns a feature representation indiscriminate between the source and target domains while still being discriminate for the original prediction task.

\begin{center}
    \includegraphics[width=8.5cm]{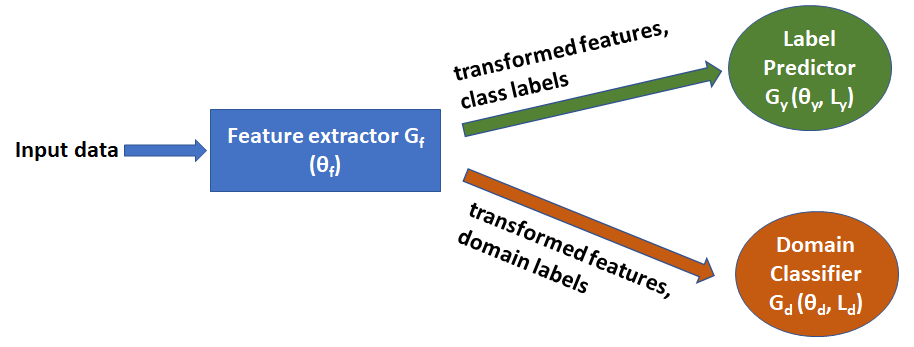}
    \captionof{figure}{DANN algorithm.}
    \label{fig:dann-workflow}
\end{center}

Formally, suppose the data is of size $N = n + n'$ where the first $n$ are from the source data and the last $n'$ are from the target data. Let $\mathcal{L}_y^i$ denote the source label prediction loss on the $i$th example and $\mathcal{L}_d^i$ denote the domain classifier loss on the $i$th example. Moreover, let $\theta_f$, $\theta_y$, and $\theta_d$ represent the parameters of $G_f$, $G_y$, and $G_d$, respectively. Then, the overall loss is given by:
\begin{equation}
\label{eq:dann_loss}
    \mathcal{L} = \frac{1}{n}\sum_{i=1}^{n}\mathcal{L}_y^i(\theta_f, \theta_y) - \lambda\left(\frac{1}{n}\sum_{i=1}^{n}\mathcal{L}_d^i(\theta_f, \theta_d) + \frac{1}{n'}\sum_{i=n+1}^{N}\mathcal{L}_d^i(\theta_f, \theta_d)\right)
\end{equation}
In particular, we choose $\mathcal{L}_y$ to be cross-entropy loss and $\mathcal{L}_d$ to be binary cross-entropy loss. The specific architectures for the networks $G_f$, $G_y$, and $G_d$ vary depending on the experiment. 
% TODO: revisit this last sentence; depends on what architectures are used for real-world data; the architecture is held fixed for the synthetic-data experiments

\subsubsection{Maximum Classifier Discrepancy}
The MCD approach (Figure \ref{fig:MCD-workflow}) considers a feature extractor $G$ and two label predictors, $F_1$ and $F_2$. The two label predictors have the same network structure and will be trained using the same training data. The two label predictors' parameters are initiated differently, which results in different optimal parameters in the final models. The MCD approach introduces the concept of \textit{discrepancy}, which is a measure of the divergence between the probabilistic outputs of $F_1$ and $F_2$. Specifically, if there are $K$ classes, suppose $p_{i,k}$ denotes the probability of the $k$th class under $p_i$, the softmax output of $F_i$, where $k \in [K]$ and $i \in \{1,2\}$. Then, the discrepancy between $p_1$ and $p_2$ is
\begin{equation}
\label{eq:mcd_discrepancy}
    d(p_1, p_2) = \frac{1}{K}\sum_{k=1}^{K}|p_{1,k} - p_{2,k}|.
\end{equation}

\begin{center}
    \includegraphics[width=8.5cm]{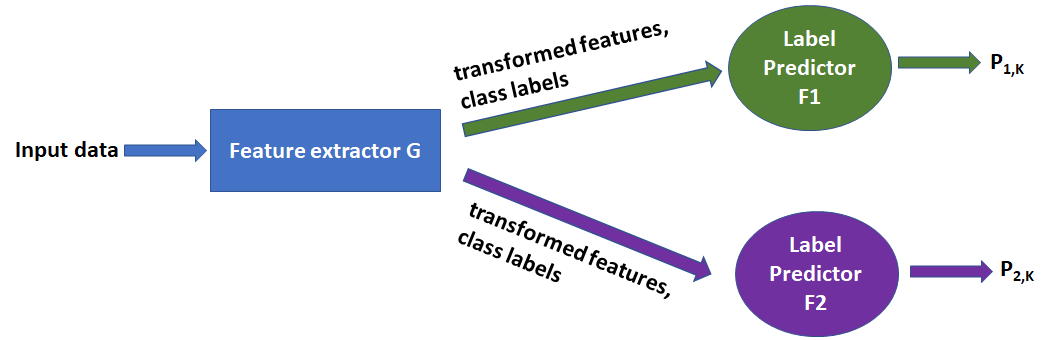}
    \captionof{figure}{MCD algorithm.}
    \label{fig:MCD-workflow}
\end{center}

A single iteration of training the overall network occurs in steps, using discrepancy to optimize for the source task while generalizing to the target task. Let $(X_s, Y_s)$ denote the source data and labels, and let $X_t$ denote the target data. To begin, the source data is fed through $G$ to $F_1$ and $F_2$, optimizing the loss on the source task
\begin{equation}
\label{eq:mcd_step_a}
    \min_{G, F_1, F_2} \mathcal{L}(X_s, Y_s),
\end{equation}
where $\mathcal{L}(X_s, Y_s)$ is given by the summed cross-entropy loss with respect to $p_1$ and $p_2$. Then $F_1$ and $F_2$ take a step to maximize the discrepancy to improve their generalization to samples outside the source support with the objective
\begin{equation}
\label{eq:mcd_step_b_1}
    \min_{F_1, F_2} \mathcal{L}(X_s, Y_s) - \lambda \mathcal{L}_{\text{adv}}(X_t)
\end{equation}
where $\mathcal{L}_{\text{adv}}(X_t) = \mathbb{E}_{\mathbf{x}_t \sim X_t} [d(p_1(y \mid \mathbf{x}_t), p_2(y \mid \mathbf{x}_t))]$.

\begin{comment}
\begin{equation}
\label{eq:mcd_step_b_2}
    \mathcal{L}_{\text{adv}}(X_t) = \mathbb{E}_{\mathbf{x}_t \sim X_t} [d(p_1(y \mid \mathbf{x}_t), p_2(y \mid \mathbf{x}_t))]
\end{equation}
\end{comment}
Just as for DANN, $\lambda$ controls the weighting of the adversarial loss $\mathcal{L}_{\text{adv}}$. Finally, $G$ takes a step to minimize discrepancy to encourage learning a representation that maps target and source data to the same space; in particular, it considers the objective
\begin{equation}
\label{eq:mcd_step_c}
    \min_G \mathcal{L}_{\text{adv}}(X_t).
\end{equation}
After training, either $F_1$ or $F_2$ (or both) can be used for inference. 
\begin{comment}
In practice, we include an additional class balance term 
\begin{equation}
\label{eq:mcd_class_balance}
    \tau \cdot \mathbb{E}_{\mathbf{x}_t \sim X_t}\left[\sum_{k=1}^{K}(\log p_1(y = k \mid \mathbf{x}_t) + \log p_2(y = k \mid \mathbf{x}_t))\right]
\end{equation}
to Equations (\ref{eq:mcd_step_a}) and (\ref{eq:mcd_step_b_1}) with $\tau$ chosen to be $0.01$, following the original paper~\cite{mcd}. 
\end{comment}
Again, the specific architectures for the networks $G$, $F_1$, and $F_2$ vary depending on the experiment.
% TODO: revisit this last sentence; depends on what architectures are used for real-world data; the architecture is held fixed for the synthetic-data experiments

\subsection{Model-Based Deep Transfer Learning}
We also consider transfer learning that adapts a source neural network model for a target setting. In particular, after training a source model, we can learn a target model (with the same architecture) by retraining parameters in three ways: (1) retrain the entire neural network, (2) freeze the first layer and retrain the remaining layers, and (3) freeze the first few layers and retrain the last layer. While each approach requires varying degrees of parameter re-tuning, all three may be feasible alternatives when the source data is not available (meaning data-based transfer learning is not applicable).

\subsection{Synthetic Experiments}
\label{subsec:methods_synthetic_experiments}
The goal of synthetic experiments is to gain insight into the conditions under which the data-based and model-based transfer learning algorithms may perform successfully in infectious disease case detection tasks. To this end, we simulate synthetic research data, of which the joint distribution is represented using Bayesian networks. Each network realization follows an architecture defined by a single class variable $z$, which represents the diagnosis, and $n$ feature variables $\{x_i\}_{i=1}^{n}$, which represent symptoms where each $x_i$ has the sole parent $z$. For notation, let $s = |\mathcal{R}(z)|$ and $r_i = |\mathcal{R}(x_i)|$ for $1 \leq i \leq n$ where $\mathcal{R}(x)$ denotes the range of $x$. In other words, $s$ and $r_i$ give the number of different diagnoses and of different symptom levels for the $i$th symptom, respectively.

Given this Bayesian network architecture, the probability tables for the target model were learned from the electronic medical records of (name of the institution) healthcare visits between January 2008 and May 2014. Training data included age groups, clinical features extracted by a medical natural language processing parser, and disease labels involving  3808 influenza visits, 6224 RSV visits, 2074 parainfluenza visits, and 2024 hMPV visits. 

Then, for the source models, we reuse the target model's marginal distribution for $z$ and generate the conditionals $p(x_i \mid z)$ as follows: We (1) convert the target model's conditional probabilities into log odds; (2) add random noise to the log odds where the noise follows a normal distribution $\mathcal{N}(0,\sigma^2)$; and (3) convert the log odds back to conditional probabilities. Specifically, we consider \\
$\sigma \in \{0.01, 0.1, 0.2, 0.3, 0.4,  0.5, 0.6, 0.7, 0.8, 0.9, 1, 1.5, 1.8, 2\}$. From these realized source models, we generate source training datasets of size 10,000, and from the target model, target training datasets of size 50, 100, 200, 400, 800, 2000, and 10,000 and a target test dataset of size 10,000. 
% did not use sizes 25 and 500

We (1) experiment assuming access to only source features/labels and target features for training to understand unsupervised data-based transfer learning and then (2) experiment assuming access to target labels for training as well to understand generally when transfer learning algorithms are applicable.

\subsubsection{Difference in Source and Target Distributions}
We explore the effect of the difference in source and target distributions on the success of the data-based transfer learning algorithms by experimenting with the synthetically simulated data. In particular, we use the KL divergence between the source distribution $p$ and target distribution $q$ as the measure of dissimilarity:
\begin{equation}
\label{eq:kl}
    KL(p, q) = \sum_{x_1, \dots, x_n, z} p(x_1, \dots, x_n, z) \log\left(\frac{p(x_1, \dots, x_n, z)}{q(x_1, \dots, x_n, z)}\right)
\end{equation}
Because our Bayesian network structure (for source and target distributions) is such that the diagnosis $z$ is the only parent for every symptom $x_i$, according to ~\cite{heckerman1995learning} we can rewrite the KL divergence as follows:
\begin{equation}
\label{eq:kl_factorized}
    KL(p, q) = \sum_{i=1}^{n} \sum_{j=1}^{s} \sum_{k=1}^{r_i} p(x_i = k \mid z = j) \cdot p(z = j) \log \left(\frac{p(x_i = k \mid z = j)}{q(x_i = k \mid z = j)}\right)
\end{equation}
Then, since $p(x_i = k \mid z = j)$ and $p(z = j)$ are given when defining the Bayesian network, Equation (\ref{eq:kl_factorized}) gives a computationally tractable form of the KL divergence. 

\subsubsection{Model Architectures for the Data-Based Algorithms}
We prefer simple architectures for faster training for the synthetic experiments; therefore, the input data has dimension $128$, and there are $4$ possible output labels. For the DANN model, we chose the feature extractor $G_f$ to be a network with two $(128 \rightarrow 128 \rightarrow 128)$ fully-connected layers with ReLU activations, the label predictor $G_y$ to be one ($128 \rightarrow 4)$ fully-connected layer with batch-norm and ReLU activation, and the domain classifier $G_d$ to be three ($128 \rightarrow 1024 \rightarrow 1024 \rightarrow 1$) fully-connected layers with batch-norm and ReLU activations for the first two layers and sigmoid activation for the final layer. Similarly, for the MCD model, we chose the feature generator $G$ to be the same as $G_f$ and the classifiers $F_1$ and $F_2$ to be three ($128 \rightarrow 128 \rightarrow 128 \rightarrow 4$) fully-connected layers, with ReLU activation for the first two layers. 

The choice for $G_d$ was borrowed from the original DANN paper~\cite{dann}, and the choices for $G_f = G$, $G_y$, $F_1$, and $F_2$ were mostly arbitrary. However, empirically, we did find that increasing the hidden dimensions to be $1024$ rather than $128$ and adding an additional fully-connected layer did not provide any meaningful performance improvement, so we stayed with the smaller network architectures. 

Our baseline models use the same architecture as the DANN model's backbone; \textit{i.e.}\ they consist of $G_f$ composed of $G_y$. When we do not assume access to the target labels, we limit the baseline model to only the source data during training; otherwise, we train the baseline on the target data.

\subsubsection{Regularizing the Adversarial Loss}
\label{subsubsec:reg_lambda_dann}
Recall from Equation (\ref{eq:dann_loss}) that the tradeoff between the domain adversarial loss \\
$\frac{1}{n}\sum_{i=1}^{n}\mathcal{L}_d^i(\theta_f, \theta_d) + \frac{1}{n'}\sum_{i=n+1}^{N}\mathcal{L}_d^i(\theta_f, \theta_d)$ and the source prediction loss $\frac{1}{n}\sum_{i=1}^{n}\mathcal{L}_y^i(\theta_f, \theta_d)$ is controlled by the coefficient $\lambda$. As a hyperparameter, $\lambda$ needs to be tuned, and by default, a standard choice is $\lambda = 1$. In the original paper introducing the DANN approach, Ganin \textit{et al.}\ consider $\lambda = 6$ in their toy experiments and use an increasing schedule from $\lambda = 0$ to $\lambda = 1$ in their benchmark experiments~\cite{dann}.

In our synthetic experiments, since we know the KL divergence between the source and target distributions (say $p$ and $q$, respectively), we tried to incorporate that information into our choice of $\lambda$. Intuitively, when $p$ and $q$ are similar (\textit{i.e.}\ $KL(p, q) \approx 0$), it is not a priority to learn a representation indiscriminate of source and target since the two come from approximately the same distribution anyway. However, when $p$ and $q$ are dissimilar (\textit{i.e.}\ $KL(p, q) \gg 0$), it is important to force the model to extract more indiscriminate features, so we prefer penalizing the domain adversarial loss more.

Na\"ively, one option is to directly choose $\lambda = KL(p, q)$. However, the KL divergence being unbounded may be undesirable, so another option is to choose
\begin{equation}
\label{eq:dann_lambda}
    \lambda = \alpha (1 - e^{- KL(p, q)})
\end{equation}
where $\alpha > 0$ are selected empirically. This chooses $\lambda \approx \alpha$ for any reasonably large $KL(p, q)$ but chooses a smaller $\lambda$ when $KL(p, q)$ is close to $0$. As an example, since in the benchmark experiments of the original DANN paper, $\lambda$ was in $[0,1]$, we may choose a setting such as $\alpha = 1$. In our experiments, we try following both the na\"ive approach and Equation (\ref{eq:dann_lambda}).

\subsection{Real-World Experiments}
We consider two real-world tasks: (1) classifying influenza positive and negative cases and (2) classifying four respiratory disease cases (Table \ref{tab:real world tl settings}). For these tasks, we tried different fully-connected networks that differ in the number of layers and units in each layer:\\
\textbf{M1:} input $\rightarrow$ 128 units (ReLU) $\rightarrow$ \# of classes (softmax) \\
\textbf{M2:} input $\rightarrow$ 64 units (ReLU) $\rightarrow$ \# of classes (softmax) \\
\textbf{M3:} input $\rightarrow$ 64 units (ReLU) $\rightarrow$ 32 units (ReLU) $\rightarrow$ \# of classes (softmax) \\
\textbf{M4:} input $\rightarrow$ 128 units (ReLU) $\rightarrow$ 128 units (ReLU) $\rightarrow$ \# of classes (softmax) \\
\textbf{M5:} input $\rightarrow$ 128 units (ReLU) $\rightarrow$ 128 units (ReLU) $\rightarrow$ 64 units (ReLU) $\rightarrow$ \# of classes (softmax) \\
\textbf{M6:} input $\rightarrow$ 128 units (ReLU) $\rightarrow$ 64 units (ReLU) $\rightarrow$ 32 units (ReLU) $\rightarrow$ 16 units (ReLU) $\rightarrow$ \# of classes (softmax)

\begin{table*}
\caption{Four real-world transfer learning settings}
  \centering
  \begin{tabular}{llll}
    \toprule
    %\multicolumn{2}{c}{Part}                   \\
    %\cmidrule{1-2}
    Task & Source & Target-train & Target-test \\
    \midrule
    flu vs. non-flu & $U_{1415}$ (670, 3276)   & $I_{14winter}$ (396, 1385) & $I_{15spring}$ (204, 3627) \\
    flu vs. non-flu &
    $I_{1415}$ (600, 5012) &  $U_{14winter}$ (455, 1311) & $U_{15spring}$ (215, 1965) \\
    flu vs. RSV vs. parainfluenza vs. hMPV &
    $U_{0815}$ (2270, 1022, 261, 307) & $I_{0814}$ (3808, 6224, 2074, 2024) & $I_{1415}$ (626, 1094, 459, 235)  \\
    flu vs. RSV vs. parainfluenza vs. hMPV &
    $I_{0815}$ (4434, 7318, 2533, 2259) & $U_{0814}$ (1828, 688, 139, 202) & $U_{1415}$ (442, 334, 122, 105) \\
    \bottomrule
  \end{tabular}
  \begin{tablenotes}\footnotesize
  \item[*]  For each healthcare system (I or U), sample sizes of each diseases are provided in parentheses.
\end{tablenotes}
  \label{tab:real world tl settings}
\end{table*}

\begin{table*}
  \caption{Comparison of six different network structures (highest value in each row is bolded)}
  \centering
  \begin{tabular}{llllllll}
    \toprule
    %\multicolumn{2}{c}{Part}                   \\
    %\cmidrule{1-2}
    Train & Test & M1 & M2 & M3 & M4 & M5 & M6\\
    \midrule
    $I_{flu-14winter}$ & $I_{flu-15spring}$ & \textbf{84.91} & 81.05 & 79.77 & 76.64  & 75.49   & 81.83\\
    $U_{flu-14winter}$ & $U_{flu-15spring}$ &  88.39 & 49.40 &  \textbf{89.22}  & 45.73  & 46.47    & 68.67\\
    $I_{4diseases-0814}$ & $I_{4diseases-1415}$ & 27.59 & 26.47 &  26.55 &  27.55    & \textbf{27.67}   & 26.01 \\
    $U_{4diseases-0814}$ & $U_{4diseases-1415}$ & 48.26 & 52.04 &  53.74 & 53.94  & \textbf{54.04}  & 45.96 \\
    \bottomrule
  \end{tabular}
  \label{tab:compare models in real world data}
\end{table*}

\begin{table*}
  \caption{Comparison of different transfer learning algorithms}
  \centering
  \begin{tabular}{lllll}
    \toprule
    %\multicolumn{2}{c}{Part}                   \\
    %\cmidrule{1-2}
    Model & $U_{flu} \rightarrow I$ & $I_{flu} \rightarrow U$ & $U_{4diseases} \rightarrow I$ & $I_{4diseases} \rightarrow U$\\
    \midrule
    network structure & M1 & M1 & M5 & M5\\
    source & 94.57 & 89.95 & 29.25 & 54.44 \\
    target & 84.91 & 88.39 & 27.67 & 54.04  \\
    MCD & (94.68) & (\textbf{90.14}) & (33.89) & (\textbf{55.73}) \\
    DANN & (\textbf{94.62}) & 87.43 & (\textbf{41.59}) & 41.67 \\
    DANN-target & 86.79 & 88.76 & (29.58) & 48.35 \\
    model-tune1 & 94.13 & 55.92 & 26.14 & 53.34 \\
    model-tune2 & 84.99 & 53.21  & 25.97 & 53.44 \\
    model-tune3 & NA & NA & 26.10 & 53.74\\
    model-tune4 & NA & NA & 26.72 & 52.94\\
    \bottomrule
  \end{tabular}
  \label{tab:compare tl models in 4 settings}
\end{table*}

We experimentally determined which of these architectures was best-suited for the two real-world tasks and used those architectures as network structures for baseline models, model-based transfer learning, and data-based transfer learning (the feature extractor).

\section{Results}
% Your experimental results. Show plots of the performance of your algorithms and interpret what they mean. Be sure to label and explain this clearly. Describe how the current results in each of the experiments align with your expectations. What metrics did you use for evaluation? How do your results compare to prior work?

\subsection{Synthetic Experiments}

\subsubsection{Choosing $\lambda$ for DANN}
\label{subsubsec:dann_lambda}
As mentioned in Section \ref{subsubsec:reg_lambda_dann}, we tried different approaches to setting the regularization parameter $\lambda$. We focused our experiments on the DANN approach since it trained faster, and we fixed our target training data size to be $400$ as a happy medium. Moreover, we considered (1) setting $\lambda \in \{1, 2, 4, 8\}$ always, (2) setting $\lambda = KL(p, q)$, and (3) setting $\lambda = \alpha (1 - e^{- KL(p,q)})$ for $\alpha \in \{1, 2, 4, 8\}$.

From Figure \ref{fig:dann_lambda}, we see that, contrary to our expectations, the choice of $\lambda$ was not that impactful on the test performance for the DANN approach. There is not a visible difference in setting $\lambda$ to a fixed constant, to the KL divergence itself, or to some (exponentially) increasing function of the KL divergence. Moreover, choosing a suitable magnitude of $\lambda$ does not appear to depend much on the dissimilarity between the source and target distributions.

\subsubsection{Varying KL Divergence and Target Data Size for DANN \& MCD}
\label{subsubsec:varying_kl_target_data_size}
Because the results of our experiments in Section \ref{subsubsec:dann_lambda} showed that the choice of $\lambda$ did not appear to be too instrumental to the model performance, for simplicity, we fixed $\lambda = 1$ for both the DANN and the MCD approaches on the synthetic data. Then we considered the effect of the difference in source and target distributions, as measured by the KL divergence given in Equation (\ref{eq:kl_factorized}) and the effect of different target training data sizes. Specifically, we considered all $14$ settings of $\sigma$ for the source models to achieve a range of KL divergences, and we tried all target training data sizes mentioned in Section \ref{subsec:methods_synthetic_experiments}.

Figures show the test performance of the DANN (Figure \ref{fig:dann_acc_vs_kl}) and MCD (Figure \ref{fig:mcd_acc_vs_kl}) approaches, respectively, compared against the baseline (source model). We see that the DANN approach achieves meaningful transfer for target training data sizes of at least $200$. The MCD approach sees slightly worse performance, but it still beats the baseline for target training sizes of at least $400$. Overall, we see that the transfer learning algorithms work to reduce the accuracy drop when working with a target distribution dissimilar from the source distribution but are unable to fully mitigate the difference. Notably though, we did not expect the MCD approach to be outclassed by the simpler DANN approach, which also took less time to train.

\subsubsection{When Is Transfer Learning Useful?}

Next, to assess the usefulness of transfer learning in general, we broadened our scope to compare the data-based transfer learning approaches with model-based transfer learning approaches and baseline methods (source model and target model) on the synthetic data. Across the different plots (Figure
\ref{fig:target50},\ref{fig:target100},\ref{fig:target200},\ref{fig:target400},\ref{fig:target800},\ref{fig:target2000},\ref{fig:target10000}), we vary the amount of target training data available.

\textbf{Target model curve vs. transfer learning model curves:} 
The target model was developed using the target training data only (\textit{i.e.}\ without source data). The target curve (in cyan) is a straight line in each subplot since we use the same target training dataset each time. We see that with a sample size of at most $800$ (\textit{i.e.}\ from Figure \ref{fig:target50},\ref{fig:target100},\ref{fig:target200},\ref{fig:target400},\ref{fig:target800}), the target model curve is lower than a transfer learning model's curve. Moreover, the benefit of transfer learning depends on the similarity between the source and target domains. When the target size is $50$, the KL divergence should be at most $3$ in order to have benefits of the source information, and when the target size is $\geq 100$, the KL divergence should be at most $0.5$. However, we emphasize that the target model requires target labels, which we may not have; in such a case, we can still use the DANN and MCD methods to outclass the baseline trained only on source data.

\textbf{Source model curve vs. transfer learning model curves:}
This comparison reiterates the results from Section \ref{subsubsec:varying_kl_target_data_size}. We refer to the baselines trained only on the source data as the source model (whose curve is in grey). Again, our experiments show that the transfer learning models perform much better than the source models when the KL divergence is greater than $1$. Moreover, the advantage is more obvious for larger divergences. Thus, we see that data-based transfer learning is very useful when the two domains are not very similar and the target domain does not have labeled training data. 

\textbf{DANN target vs. DANN vs. MCD:} When the KL divergence is greater than 3, the DANN approach with access to the target labels performs significantly better than the normal DANN approach, and the normal DANN approach performs better than the MCD approach.

\textbf{Model-tune3 vs. model-tune2 vs. model-tune1:} The numeric suffix denotes the number of layers tuned (following the model-based approach). We see that when the target training data size is sufficient, tuning more layers works better, especially if the KL divergence is large. 

\textbf{Model-based vs data-based:} When the KL divergence is large, the fully-tuned models perform better than the data-based models on the synthetic data. When the KL divergence is small and target training data is insufficient, data-based models perform similarly to model-based models.

\subsection{Real-World Experiments}
\subsubsection{Model Architecture Search}
Table \ref{tab:compare models in real world data} shows that, for the flu classification tasks (rows 1-2), a simple one-hidden layer (128 units) network, M1, performs relatively well. This indicates that the flu classification task only needs a simple model. In contrast, for the four-disease classification tasks (rows 3-4), a three-layer network, M5, performs the best, indicating that a deeper network may be better for more complicated tasks. Thus, we use M1 for further experiments related to the flu classification tasks and M5 for further experiments related to the four-disease classification tasks in Section \ref{subsubsec:models for real world}.

\subsubsection{Compare transfer learning algorithms}
\label{subsubsec:models for real world}

For the four settings in Table \ref{tab:real world tl settings}, Table \ref{tab:compare tl models in 4 settings} lists the performance of the different models for the four settings in Table 1. The highest value per row is bolded and model accuracy values greater than those of both the source and target models are in parentheses. Interestingly, the source models all performed better than the target models. The source training datasets combine the source institution's visits in historical seasons and the most recent season, while the target training datasets contain the target institution's visits in historical seasons and have a smaller sample size in all settings except the third.

Results for this experiment show that transfer learning may be useful for infectious disease classification tasks. Although the accuracy gains for transfer learning in setting 1, 2, and 4 were not significant, in the third setting, transfer learning models clearly outperformed just training on the source data or target data. In all four of the settings, the data-based transfer learning models (MCD or DANN) always performed the best (bold in each column). The MCD model performed the best when sharing from I to U, and the DANN model performed the best when sharing from U to I.

Contrary to expectations, tuning the models (the last four rows) resulted in a performance worse than that of the source models. For settings 3 and 4, one possible explanation is that there is a relatively significant domain shift in the joint distribution $P(\text{findings}, \text{disease})$ from the historical seasons (target training data) to the most recent season (target test data). The domain shift in the marginal $P(\text{findings})$ may be relatively small, allowing data-based methods use the unlabeled target data to perform better than the source models.

For settings 1 and 2, unfortunately, we do not have any good explanation as to why using the target training data to tune the source models resulted in reduced performance (especially the dramatic drops in setting 2). In fact, in these two settings, the target training dataset and the test dataset were from the first half and the second half of one season in the same institute respectively and thus were expected to have much less domain shift.

\section{Discussion}
\subsection{Synthetic Experiments}
Both of the data-based transfer learning algorithms that we explored, DANN and MCD, were originally proposed in the context of images~\cite{dann}\cite{mcd}, while here, we applied the techniques to infectious disease classification. Notably, the features considered in the synthetic experiments were all categorical corresponding to symptoms as opposed to the spatially-structured pixels that comprise an image. To accommodate this difference, we replaced convolutional layers with fully-connected layers in the suggested architectures. Perhaps, this substitution hindered the performance of the transfer-learning approaches. While we experimented with one-step changes such as increasing the hidden dimensions or adding additional layers, we did not thoroughly investigate more sophisticated changes such as incorporating varying levels of dropout or employing more complex layer architectures. 

Still, since the goal of the synthetic experiments was to understand general trends and relationships, the absolute test accuracy was not the focus. Rather, from our exploration, we found that the coefficient $\lambda$ controlling the relative weighting of the adversarial loss was not impactful in the DANN algorithm, suggesting that there may not be a need to tune it precisely. Furthermore, we evidenced that transfer learning does have its place in this infectious disease context, as it was able to reduce the performance drop when predicting on increasingly dissimilar target domains. This observation partially motivated our further investigation on the real-world data.

We also saw success with the model-based approach. In particular, we saw that tuning more layers was important when the KL divergence between the source and target distributions was large. Hence, in general in the synthetic setting, we learned that we should not limit ourselves to only retraining one or two layers if possible. As such, we can view the entire source model simply as a warm start that provides a more favorable initialization for the model meant for the target task. 

\subsection{Real-World Data Experiments}
Our real-world data experiments demonstrated that data-based transfer learning, especially the MCD approach, may develop a model that performs better than both a source model and a target model. However, model-based transfer learning did not further improve the performance of a source model that was set as a warm start. We hypothesize that this result is due to the domain shift in the joint distribution of clinical findings and disease between the target training data and the target test data. As mentioned earlier, the training data was older, while the test data came from the most recent season. This shift did not exist for the synthetic experiments since in that case the training and test data were randomly generated from the same distribution, so we did not see such a drop in performance. Hence, in the future, we must also account for temporal differences in the training and test data as a part of the domain shift. In other words, if we apply a transfer learning method, we should consider the shift between source and target domains as well as within the target domain over time. 

\section{Conclusion}
To conclude, in the context of infectious disease classification, transfer learning may be useful when (1) the source and target are similar and the target training data is insufficient and (2) the target training data does not have labels. We saw that model-based transfer learning works well in the first situation, where case the performance closely matched that of the data-based transfer learning models. However, further investigation is needed to understand the drop in performance that model-based transfer learning experiences from the domain shift in real world research.

\begin{center}
    \includegraphics[width=9cm]{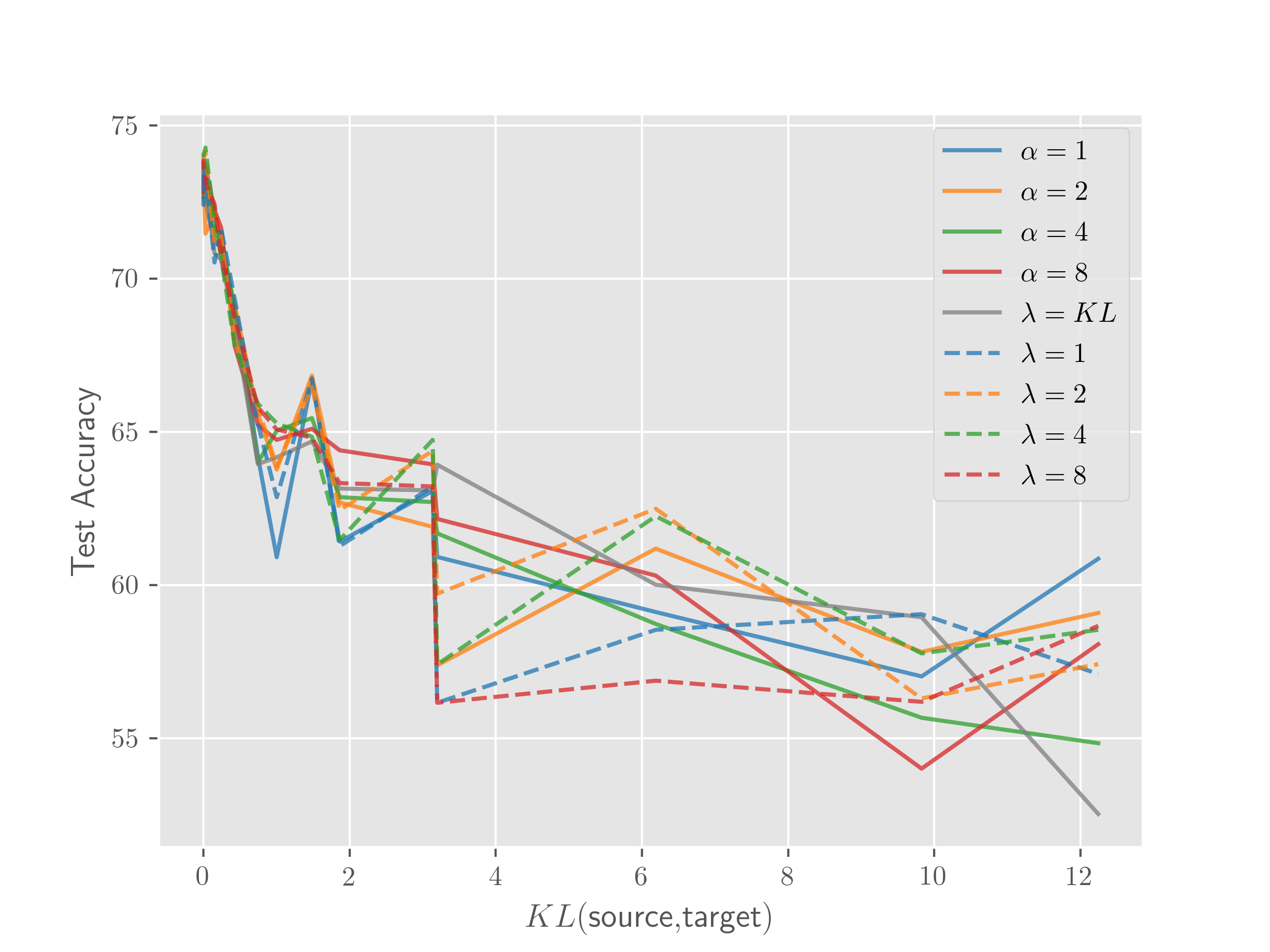}
    \captionof{figure}{DANN test accuracy with different settings of $\lambda$.}
    \label{fig:dann_lambda}
\end{center}

\begin{center}
    \includegraphics[width=9cm]{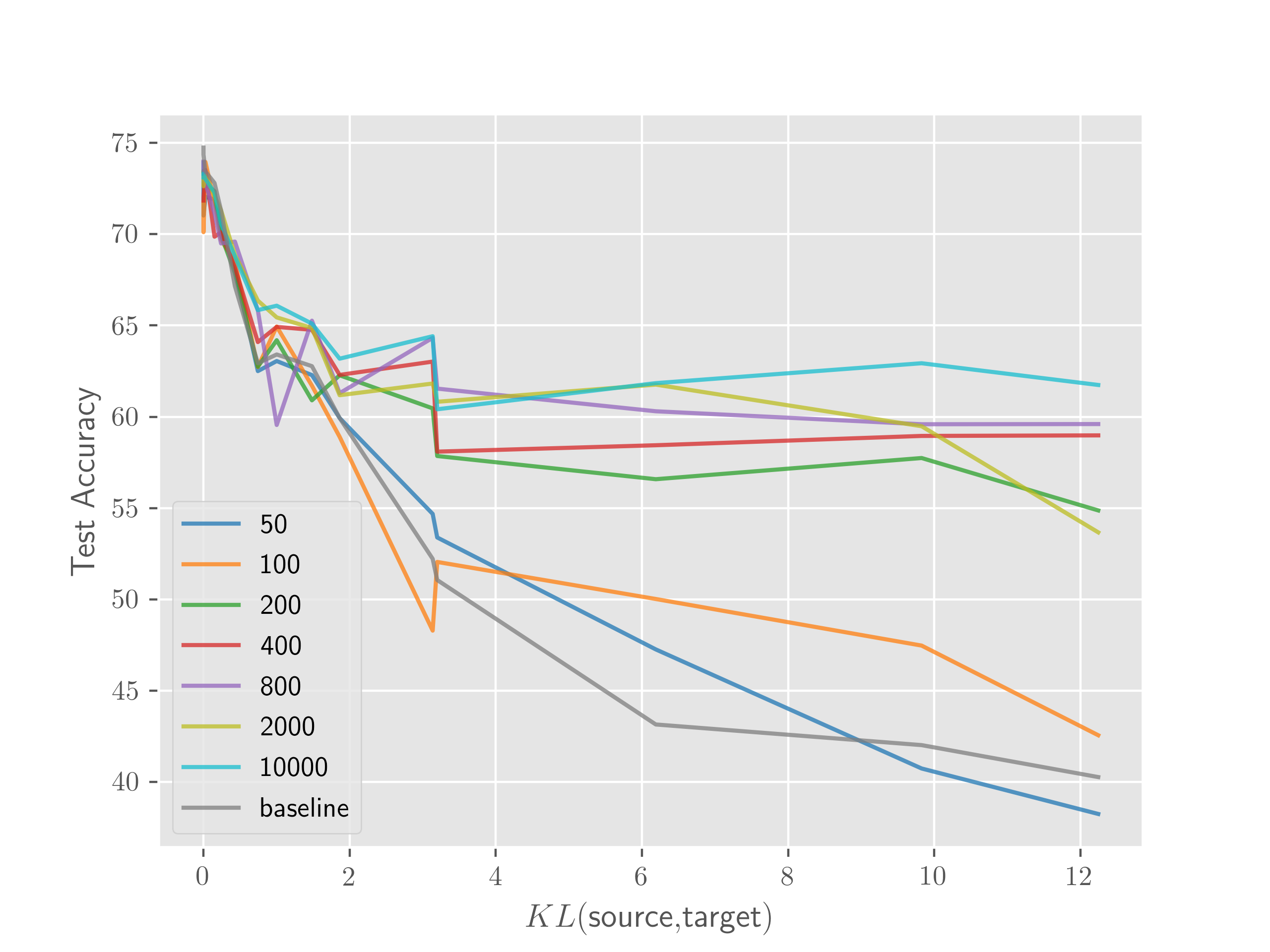}
    \captionof{figure}{DANN: test accuracy with different target training data sizes.}
    \label{fig:dann_acc_vs_kl}
\end{center}

\begin{center}
    \includegraphics[width=9cm]{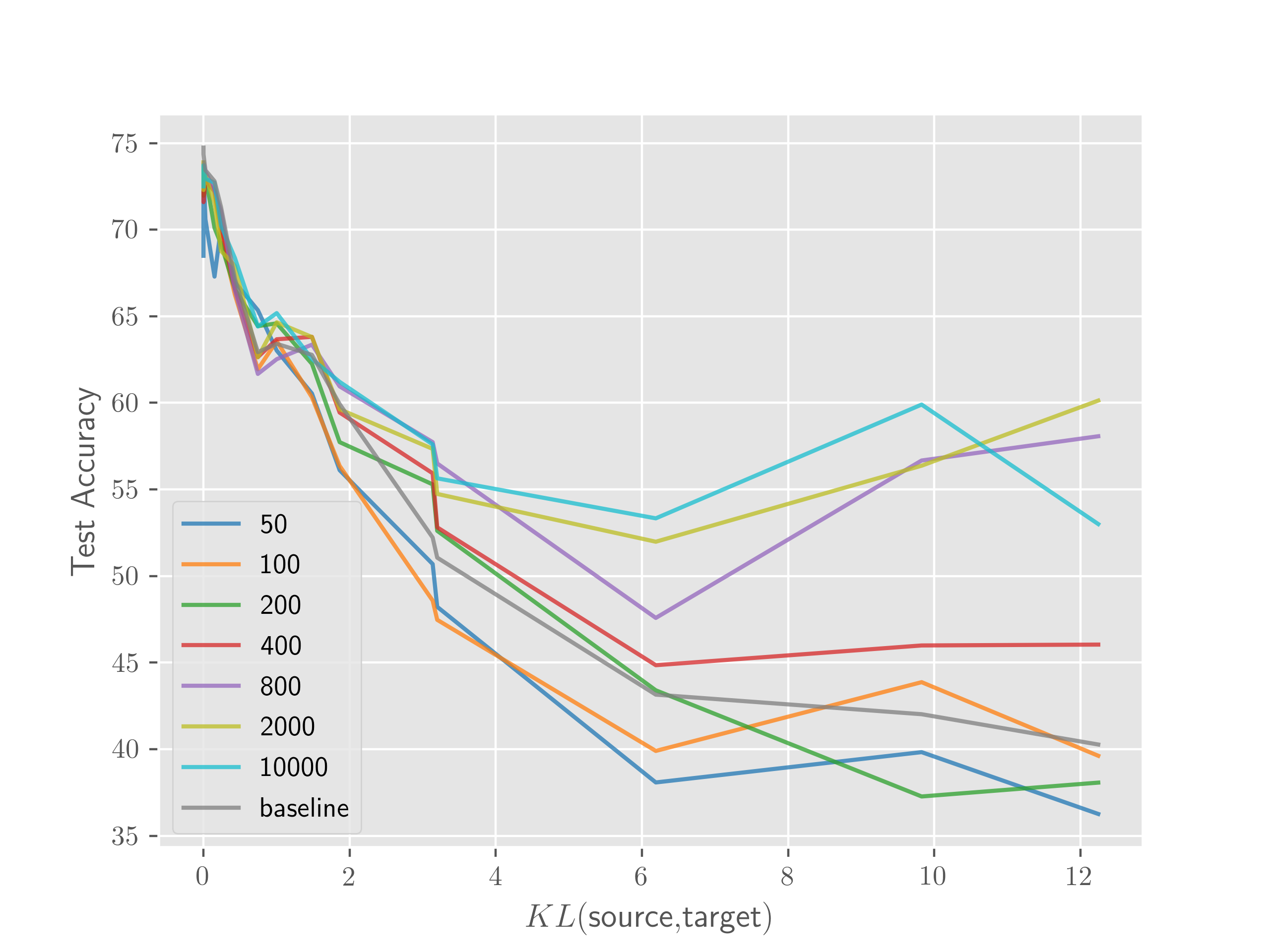} 
    \captionof{figure}{MCD: test accuracy with different target training data sizes.}
    \label{fig:mcd_acc_vs_kl}
\end{center}

\begin{center}
    \includegraphics[width=9cm]{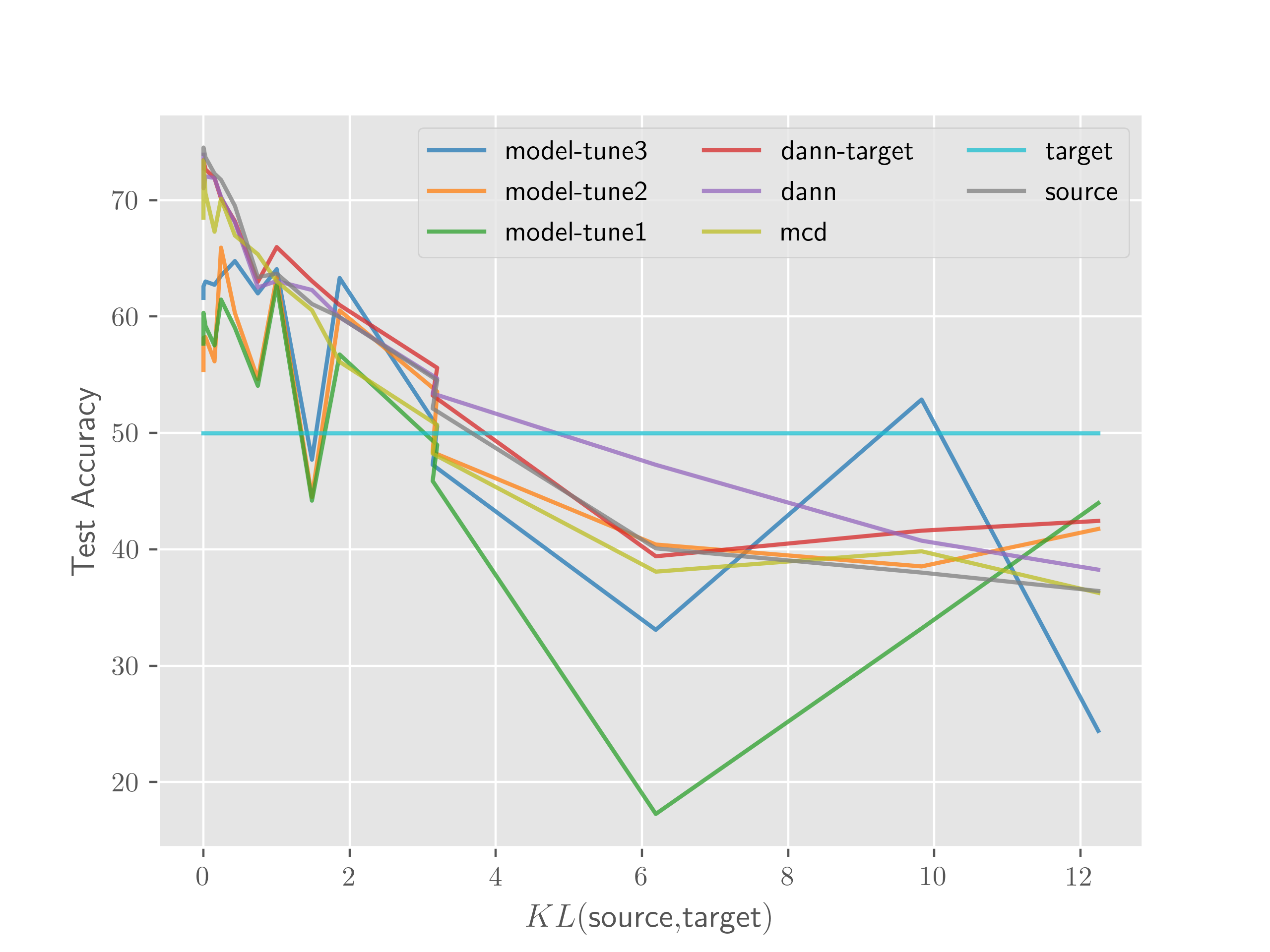} 
    \captionof{figure}{Comparisons of model performance (target sample size = 50).}
    \label{fig:target50}
\end{center}

\begin{center}
    \includegraphics[width=9cm]{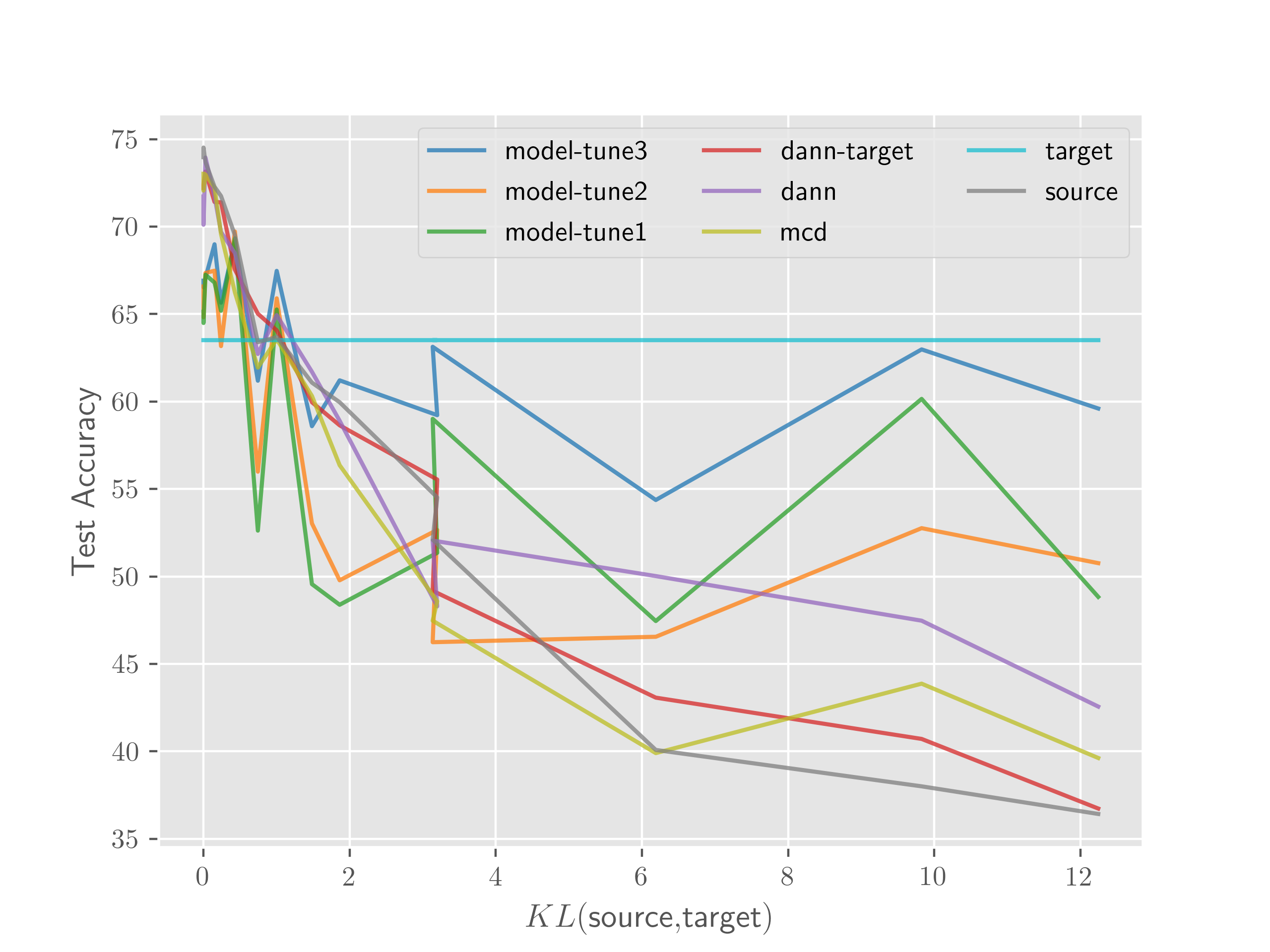}
    \captionof{figure}{Comparisons of model performance (target sample size = 100).}
    \label{fig:target100}
\end{center}

\begin{center}
    \includegraphics[width=9cm]{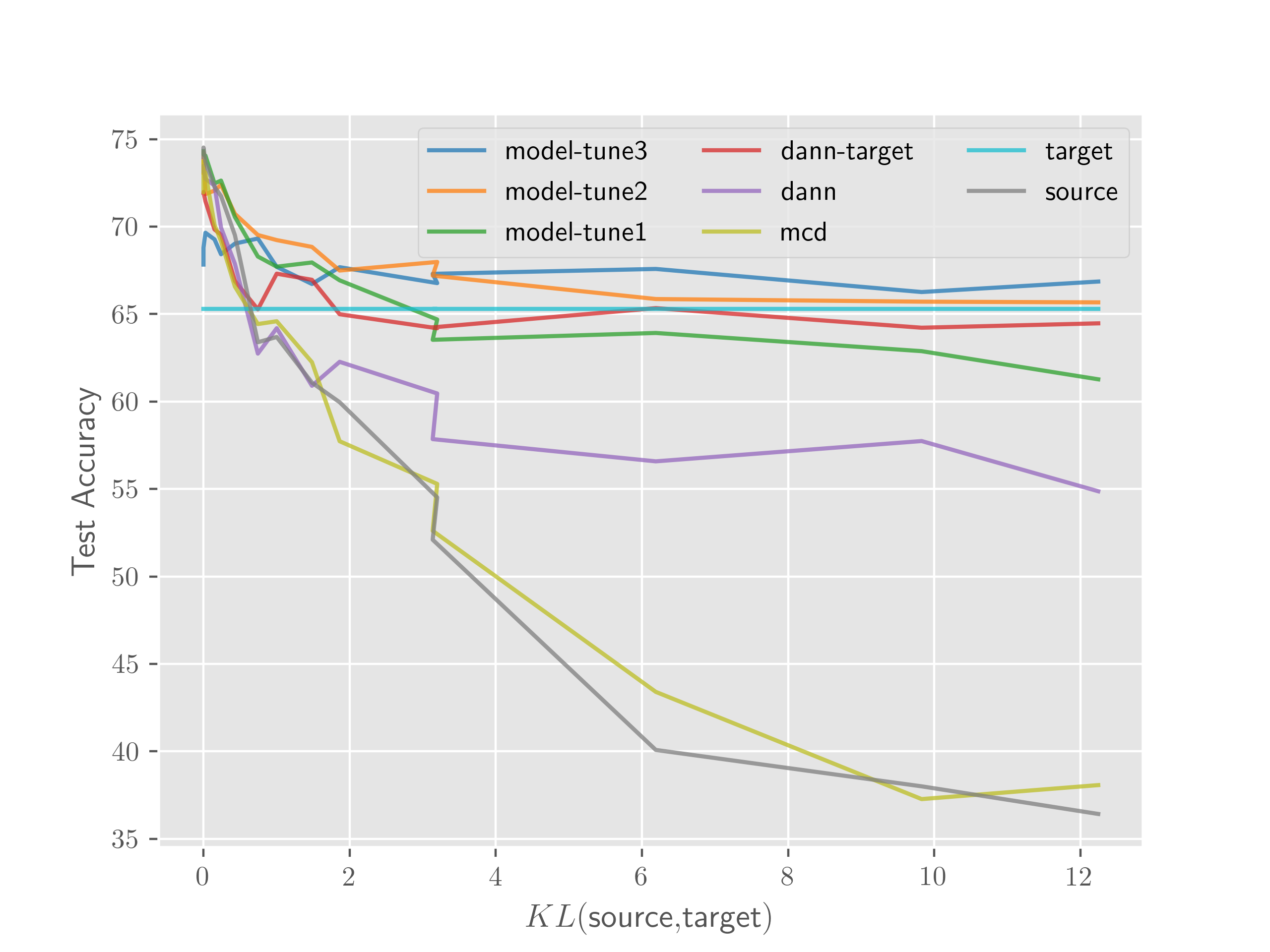}
    \captionof{figure}{Comparisons of model performance (target sample size = 200).}
    \label{fig:target200}
\end{center}

\begin{center}
    \includegraphics[width=9cm]{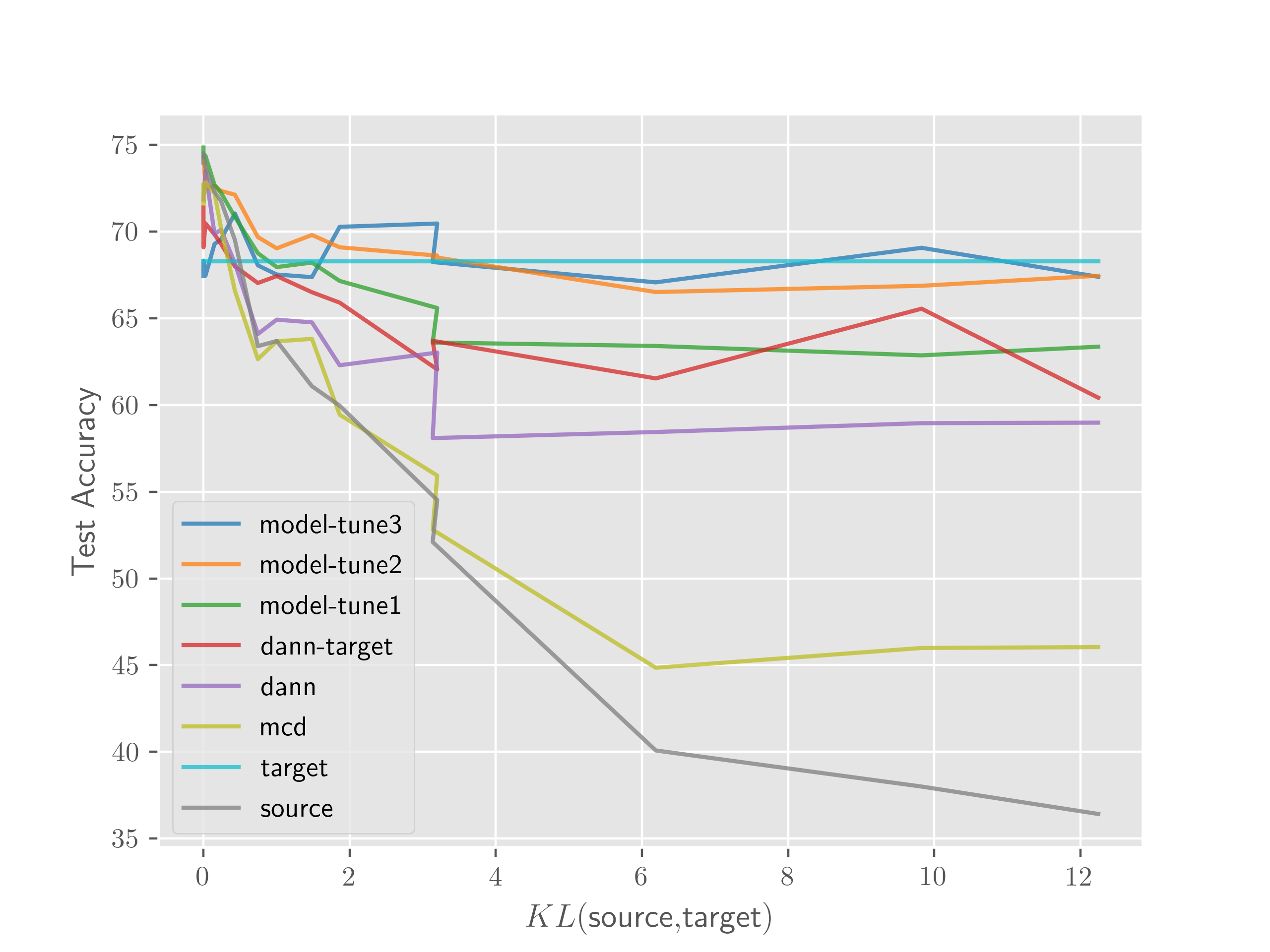}
    \captionof{figure}{Comparisons of model performance (target sample size = 400).}
    \label{fig:target400}
\end{center}

\begin{center}
    \includegraphics[width=9cm]{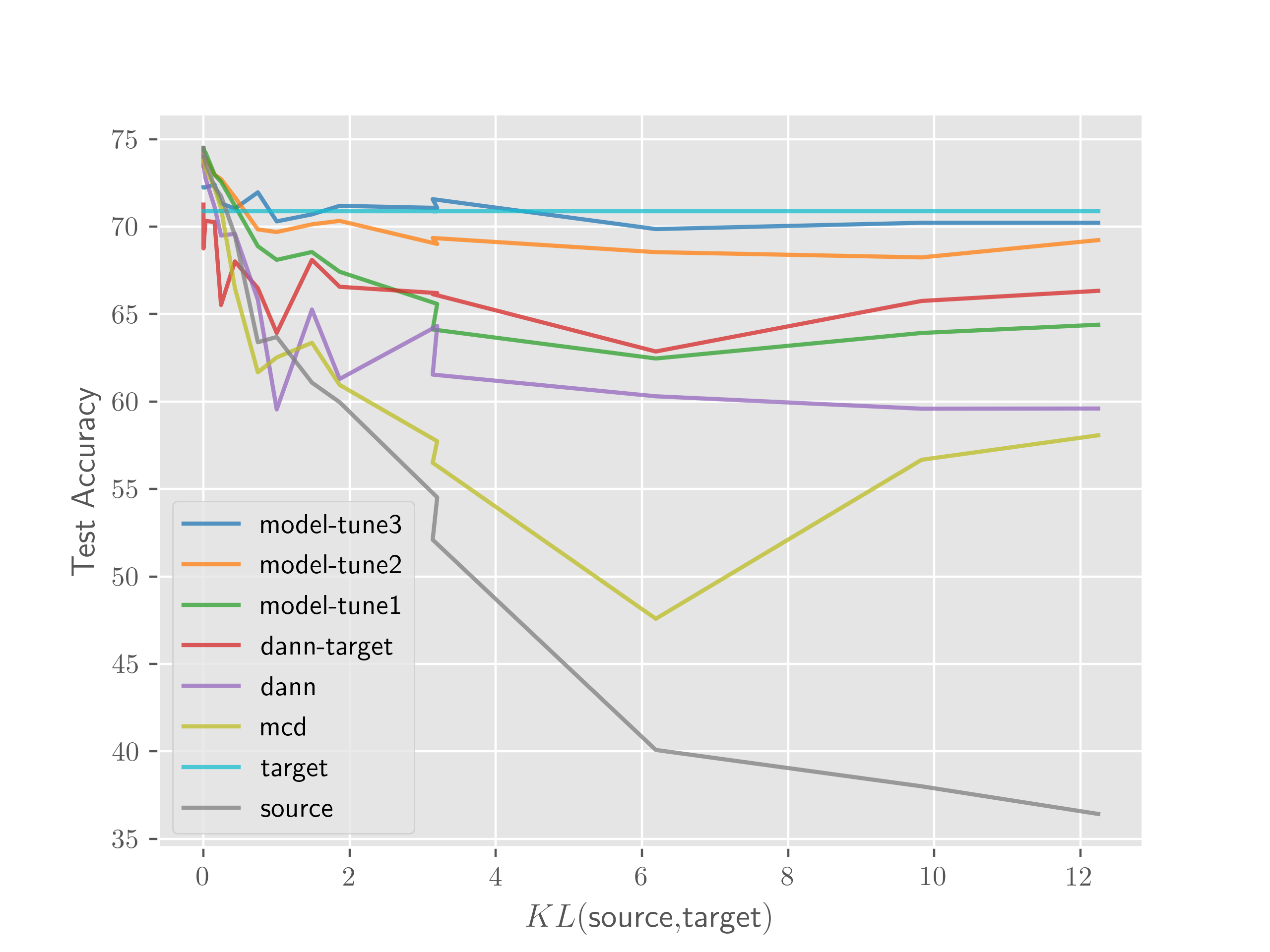}
    \captionof{figure}{Comparisons of model performance (target sample size = 800).}
    \label{fig:target800}
\end{center}

\begin{center}
    \includegraphics[width=9cm]{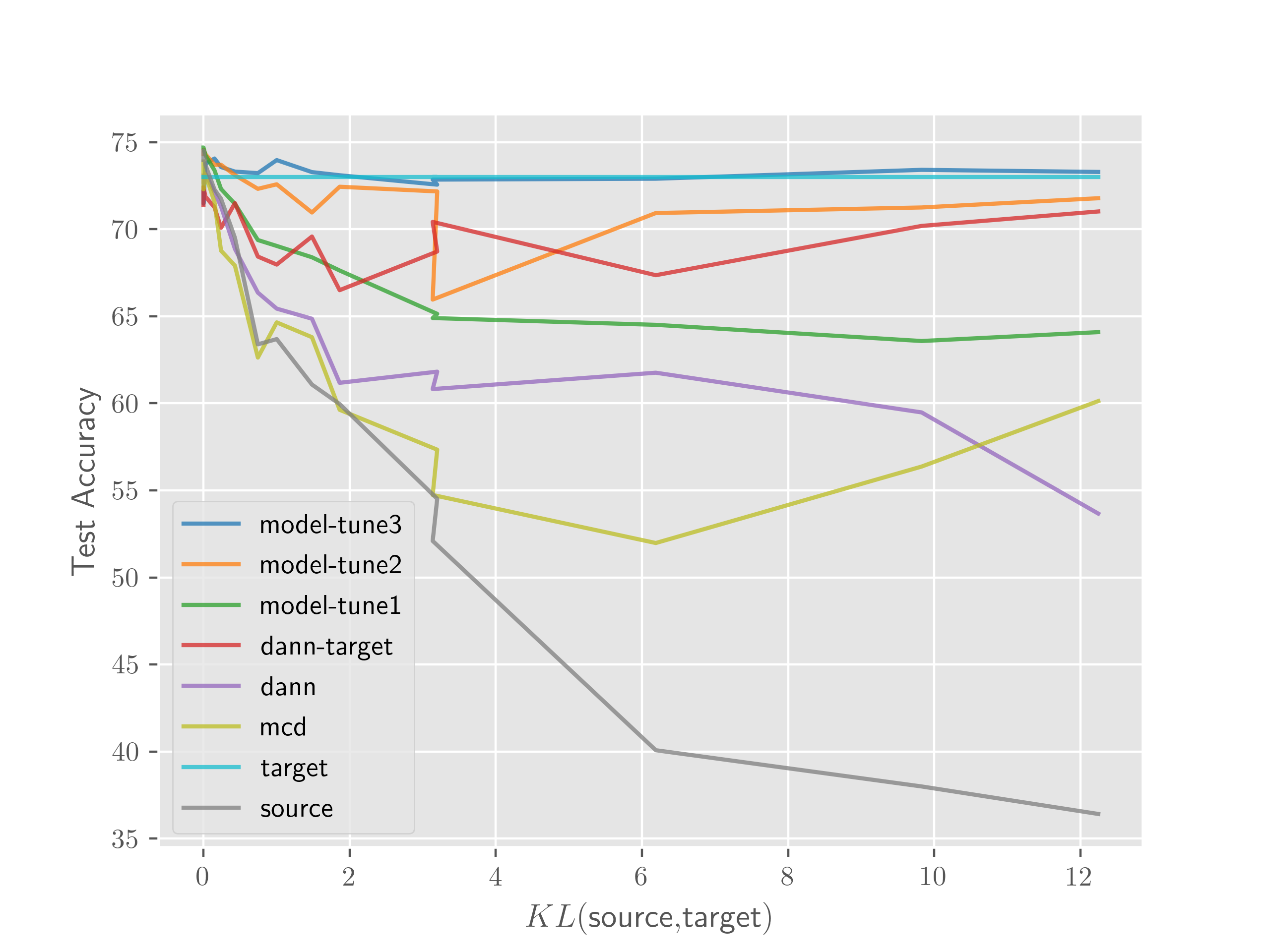}
    \captionof{figure}{Comparisons of model performance (target sample size = 2000).}
    \label{fig:target2000}
\end{center}

\begin{center}
    \includegraphics[width=9cm]{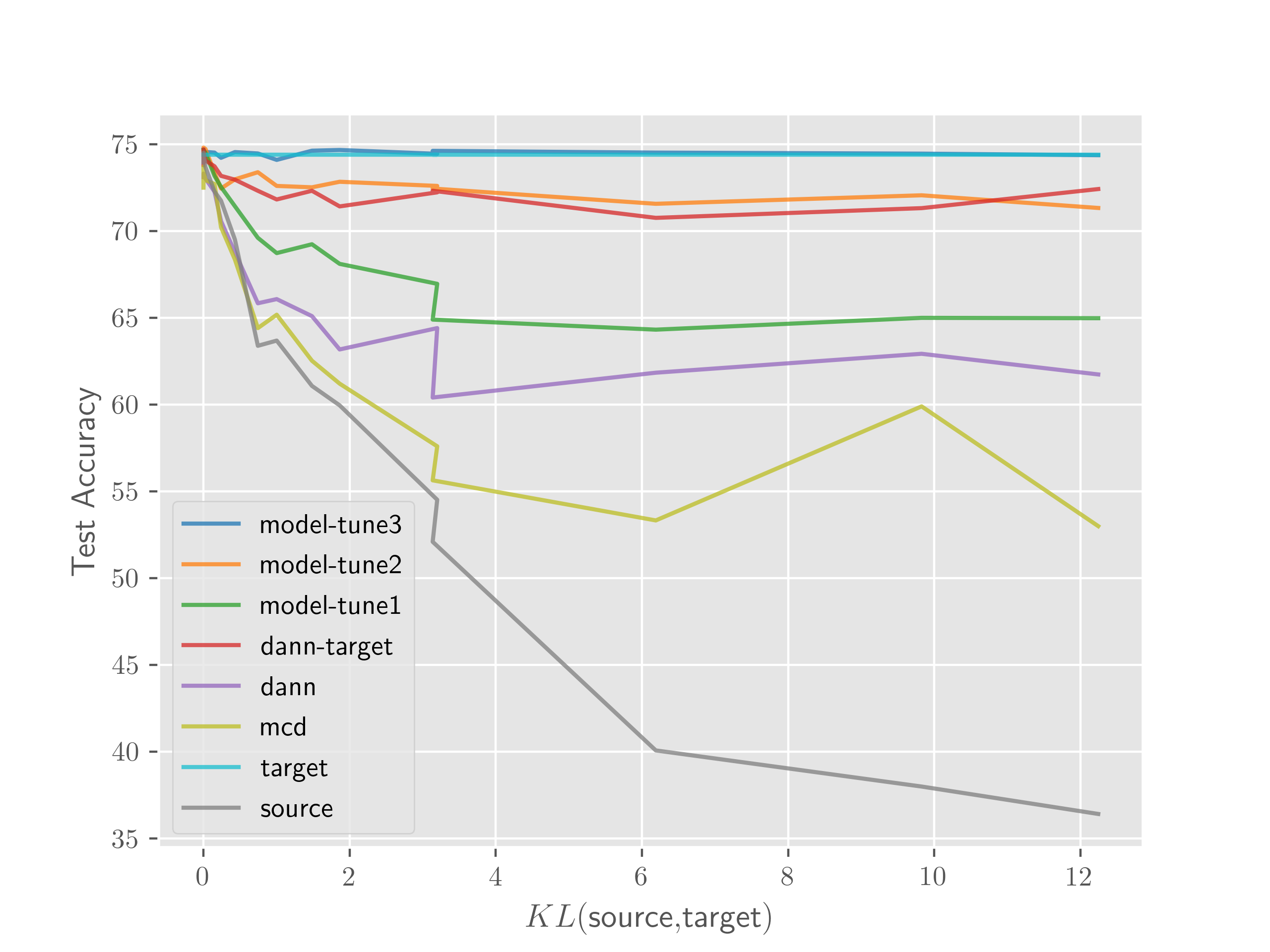}
    \captionof{figure}{Comparisons of model performance (target sample size = 10000).}
    \label{fig:target10000}
\end{center}

\section*{IRB Information}
The research data was collected through PRO13030471 and study LM011370.

\section*{Acknowledgments}
Dr. Ye Ye was supported by NIH NLM K99LM013383 grant, under the guidance of Drs. Michael Becich, Gregory Cooper, Michael Wagner, and Heng Huang. We would like to thank our collaborators, Drs. Jeffery Ferraro, Peter Haug, and Per Gesteland from Intermountain Healthcare, for providing research data. We would like to thank Jonathon Byrd and Drs. Ruslan Salakhutdinov, Andrej Risteski for research advice. Starter code for the data-based and model-based transfer learning algorithms was taken from~\cite{dalib}.

\bibliographystyle{ACM-Reference-Format}
\bibliography{reference}


\begin{thebibliography}{22}

%%% the Web address.

\ifx \showCODEN    \undefined \def \showCODEN     #1{\unskip}     \fi
\ifx \showDOI      \undefined \def \showDOI       #1{#1}\fi
\ifx \showISBNx    \undefined \def \showISBNx     #1{\unskip}     \fi
\ifx \showISBNxiii \undefined \def \showISBNxiii  #1{\unskip}     \fi
\ifx \showISSN     \undefined \def \showISSN      #1{\unskip}     \fi
\ifx \showLCCN     \undefined \def \showLCCN      #1{\unskip}     \fi
\ifx \shownote     \undefined \def \shownote      #1{#1}          \fi
\ifx \showarticletitle \undefined \def \showarticletitle #1{#1}   \fi
\ifx \showURL      \undefined \def \showURL       {\relax}        \fi
% invisible to TeX
\providecommand\bibfield[2]{#2}
\providecommand\bibinfo[2]{#2}
\providecommand\natexlab[1]{#1}
\providecommand\showeprint[2][]{arXiv:#2}

\bibitem[\protect\citeauthoryear{Bickel, Br{\"u}ckner, and Scheffer}{Bickel
  et~al\mbox{.}}{2007}]%
        {bickel2007discriminative}
\bibfield{author}{\bibinfo{person}{Steffen Bickel}, \bibinfo{person}{Michael
  Br{\"u}ckner}, {and} \bibinfo{person}{Tobias Scheffer}.}
  \bibinfo{year}{2007}\natexlab{}.
\newblock \showarticletitle{Discriminative learning for differing training and
  test distributions}. In \bibinfo{booktitle}{\emph{Proceedings of the 24th
  international conference on Machine learning}}. \bibinfo{pages}{81--88}.
\newblock


\bibitem[\protect\citeauthoryear{Dai, Xue, Yang, and Yu}{Dai
  et~al\mbox{.}}{2007a}]%
        {dai2007transferring}
\bibfield{author}{\bibinfo{person}{Wenyuan Dai}, \bibinfo{person}{Gui-Rong
  Xue}, \bibinfo{person}{Qiang Yang}, {and} \bibinfo{person}{Yong Yu}.}
  \bibinfo{year}{2007}\natexlab{a}.
\newblock \showarticletitle{Transferring naive bayes classifiers for text
  classification}. In \bibinfo{booktitle}{\emph{AAAI}},
  Vol.~\bibinfo{volume}{7}. \bibinfo{pages}{540--545}.
\newblock


\bibitem[\protect\citeauthoryear{Dai, Yang, Xue, and Yu}{Dai
  et~al\mbox{.}}{2007b}]%
        {tradaboost}
\bibfield{author}{\bibinfo{person}{Wenyuan Dai}, \bibinfo{person}{Qiang Yang},
  \bibinfo{person}{Gui-Rong Xue}, {and} \bibinfo{person}{Yong Yu}.}
  \bibinfo{year}{2007}\natexlab{b}.
\newblock \showarticletitle{Boosting for Transfer Learning}. In
  \bibinfo{booktitle}{\emph{Proceedings of the 24th International Conference on
  Machine Learning}} (Corvalis, Oregon, USA) \emph{(\bibinfo{series}{ICML
  '07})}. \bibinfo{publisher}{Association for Computing Machinery},
  \bibinfo{address}{New York, NY, USA}, \bibinfo{pages}{193–200}.
\newblock
\showISBNx{9781595937933}
\urldef\tempurl%
\url{https://doi.org/10.1145/1273496.1273521}
\showDOI{\tempurl}


\bibitem[\protect\citeauthoryear{Evgeniou and Pontil}{Evgeniou and
  Pontil}{2004}]%
        {evgeniou2004regularized}
\bibfield{author}{\bibinfo{person}{Theodoros Evgeniou} {and}
  \bibinfo{person}{Massimiliano Pontil}.} \bibinfo{year}{2004}\natexlab{}.
\newblock \showarticletitle{Regularized multi--task learning}. In
  \bibinfo{booktitle}{\emph{Proceedings of the tenth ACM SIGKDD international
  conference on Knowledge discovery and data mining}}.
  \bibinfo{pages}{109--117}.
\newblock


\bibitem[\protect\citeauthoryear{Ganin, Ustinova, Ajakan, Germain, Larochelle,
  Laviolette, Marchand, and Lempitsky}{Ganin et~al\mbox{.}}{2016}]%
        {dann}
\bibfield{author}{\bibinfo{person}{Yaroslav Ganin}, \bibinfo{person}{Evgeniya
  Ustinova}, \bibinfo{person}{Hana Ajakan}, \bibinfo{person}{Pascal Germain},
  \bibinfo{person}{Hugo Larochelle}, \bibinfo{person}{François Laviolette},
  \bibinfo{person}{Mario Marchand}, {and} \bibinfo{person}{Victor Lempitsky}.}
  \bibinfo{year}{2016}\natexlab{}.
\newblock \bibinfo{title}{Domain-Adversarial Training of Neural Networks}.
\newblock
\newblock
\showeprint[arxiv]{1505.07818}~[stat.ML]


\bibitem[\protect\citeauthoryear{Heckerman, Geiger, and Chickering}{Heckerman
  et~al\mbox{.}}{1995}]%
        {heckerman1995learning}
\bibfield{author}{\bibinfo{person}{David Heckerman}, \bibinfo{person}{Dan
  Geiger}, {and} \bibinfo{person}{David~M Chickering}.}
  \bibinfo{year}{1995}\natexlab{}.
\newblock \showarticletitle{Learning Bayesian networks: The combination of
  knowledge and statistical data}.
\newblock \bibinfo{journal}{\emph{Machine learning}} \bibinfo{volume}{20},
  \bibinfo{number}{3} (\bibinfo{year}{1995}), \bibinfo{pages}{197--243}.
\newblock


\bibitem[\protect\citeauthoryear{Jebara}{Jebara}{2004}]%
        {jebara2004multi}
\bibfield{author}{\bibinfo{person}{Tony Jebara}.}
  \bibinfo{year}{2004}\natexlab{}.
\newblock \showarticletitle{Multi-task feature and kernel selection for SVMs}.
  In \bibinfo{booktitle}{\emph{Proceedings of the twenty-first international
  conference on Machine learning}}. \bibinfo{pages}{55}.
\newblock


\bibitem[\protect\citeauthoryear{Junguang~Jiang}{Junguang~Jiang}{2020}]%
        {dalib}
\bibfield{author}{\bibinfo{person}{Mingsheng~Long Junguang~Jiang, Bo~Fu}.}
  \bibinfo{year}{2020}\natexlab{}.
\newblock \bibinfo{title}{Transfer-Learning-library}.
\newblock
  \bibinfo{howpublished}{\url{https://github.com/thuml/Transfer-Learning-Library}}.
\newblock


\bibitem[\protect\citeauthoryear{Kermany, Goldbaum, Cai, Valentim, Liang,
  Baxter, McKeown, Yang, Wu, Yan, et~al\mbox{.}}{Kermany et~al\mbox{.}}{2018}]%
        {kermany2018identifying}
\bibfield{author}{\bibinfo{person}{Daniel~S Kermany}, \bibinfo{person}{Michael
  Goldbaum}, \bibinfo{person}{Wenjia Cai}, \bibinfo{person}{Carolina~CS
  Valentim}, \bibinfo{person}{Huiying Liang}, \bibinfo{person}{Sally~L Baxter},
  \bibinfo{person}{Alex McKeown}, \bibinfo{person}{Ge Yang},
  \bibinfo{person}{Xiaokang Wu}, \bibinfo{person}{Fangbing Yan},
  {et~al\mbox{.}}} \bibinfo{year}{2018}\natexlab{}.
\newblock \showarticletitle{Identifying medical diagnoses and treatable
  diseases by image-based deep learning}.
\newblock \bibinfo{journal}{\emph{Cell}} \bibinfo{volume}{172},
  \bibinfo{number}{5} (\bibinfo{year}{2018}), \bibinfo{pages}{1122--1131}.
\newblock


\bibitem[\protect\citeauthoryear{Kim, Wang, Zhao, Im, Min, Choi, Tadros, Choi,
  Castro, Weissleder, et~al\mbox{.}}{Kim et~al\mbox{.}}{2018}]%
        {kim2018deep}
\bibfield{author}{\bibinfo{person}{Sung-Jin Kim}, \bibinfo{person}{Chuangqi
  Wang}, \bibinfo{person}{Bing Zhao}, \bibinfo{person}{Hyungsoon Im},
  \bibinfo{person}{Jouha Min}, \bibinfo{person}{Hee~June Choi},
  \bibinfo{person}{Joseph Tadros}, \bibinfo{person}{Nu~Ri Choi},
  \bibinfo{person}{Cesar~M Castro}, \bibinfo{person}{Ralph Weissleder},
  {et~al\mbox{.}}} \bibinfo{year}{2018}\natexlab{}.
\newblock \showarticletitle{Deep transfer learning-based hologram
  classification for molecular diagnostics}.
\newblock \bibinfo{journal}{\emph{Scientific reports}} \bibinfo{volume}{8},
  \bibinfo{number}{1} (\bibinfo{year}{2018}), \bibinfo{pages}{1--12}.
\newblock


\bibitem[\protect\citeauthoryear{Liao, Xue, and Carin}{Liao
  et~al\mbox{.}}{2005}]%
        {liao2005logistic}
\bibfield{author}{\bibinfo{person}{Xuejun Liao}, \bibinfo{person}{Ya Xue},
  {and} \bibinfo{person}{Lawrence Carin}.} \bibinfo{year}{2005}\natexlab{}.
\newblock \showarticletitle{Logistic regression with an auxiliary data source}.
  In \bibinfo{booktitle}{\emph{Proceedings of the 22nd international conference
  on Machine learning}}. \bibinfo{pages}{505--512}.
\newblock


\bibitem[\protect\citeauthoryear{Mazo, Bernal, Trujillo, and Alegre}{Mazo
  et~al\mbox{.}}{2018}]%
        {mazo2018transfer}
\bibfield{author}{\bibinfo{person}{Claudia Mazo}, \bibinfo{person}{Jose
  Bernal}, \bibinfo{person}{Maria Trujillo}, {and} \bibinfo{person}{Enrique
  Alegre}.} \bibinfo{year}{2018}\natexlab{}.
\newblock \showarticletitle{Transfer learning for classification of
  cardiovascular tissues in histological images}.
\newblock \bibinfo{journal}{\emph{Computer methods and programs in
  biomedicine}}  \bibinfo{volume}{165} (\bibinfo{year}{2018}),
  \bibinfo{pages}{69--76}.
\newblock


\bibitem[\protect\citeauthoryear{Pan, Tsang, Kwok, and Yang}{Pan
  et~al\mbox{.}}{2010}]%
        {pan2010domain}
\bibfield{author}{\bibinfo{person}{Sinno~Jialin Pan}, \bibinfo{person}{Ivor~W
  Tsang}, \bibinfo{person}{James~T Kwok}, {and} \bibinfo{person}{Qiang Yang}.}
  \bibinfo{year}{2010}\natexlab{}.
\newblock \showarticletitle{Domain adaptation via transfer component analysis}.
\newblock \bibinfo{journal}{\emph{IEEE Transactions on Neural Networks}}
  \bibinfo{volume}{22}, \bibinfo{number}{2} (\bibinfo{year}{2010}),
  \bibinfo{pages}{199--210}.
\newblock


\bibitem[\protect\citeauthoryear{Pan and Yang}{Pan and Yang}{2009}]%
        {pan2009survey}
\bibfield{author}{\bibinfo{person}{Sinno~Jialin Pan} {and}
  \bibinfo{person}{Qiang Yang}.} \bibinfo{year}{2009}\natexlab{}.
\newblock \showarticletitle{A survey on transfer learning}.
\newblock \bibinfo{journal}{\emph{IEEE Transactions on knowledge and data
  engineering}} \bibinfo{volume}{22}, \bibinfo{number}{10}
  (\bibinfo{year}{2009}), \bibinfo{pages}{1345--1359}.
\newblock


\bibitem[\protect\citeauthoryear{Ribeiro, Uhl, Wimmer, and H{\"a}fner}{Ribeiro
  et~al\mbox{.}}{2016}]%
        {ribeiro2016exploring}
\bibfield{author}{\bibinfo{person}{Eduardo Ribeiro}, \bibinfo{person}{Andreas
  Uhl}, \bibinfo{person}{Georg Wimmer}, {and} \bibinfo{person}{Michael
  H{\"a}fner}.} \bibinfo{year}{2016}\natexlab{}.
\newblock \showarticletitle{Exploring deep learning and transfer learning for
  colonic polyp classification}.
\newblock \bibinfo{journal}{\emph{Computational and mathematical methods in
  medicine}}  \bibinfo{volume}{2016} (\bibinfo{year}{2016}).
\newblock


\bibitem[\protect\citeauthoryear{Sachan, Xie, Sachan, and Xing}{Sachan
  et~al\mbox{.}}{2018}]%
        {sachan2018effective}
\bibfield{author}{\bibinfo{person}{Devendra~Singh Sachan},
  \bibinfo{person}{Pengtao Xie}, \bibinfo{person}{Mrinmaya Sachan}, {and}
  \bibinfo{person}{Eric~P Xing}.} \bibinfo{year}{2018}\natexlab{}.
\newblock \showarticletitle{Effective use of bidirectional language modeling
  for transfer learning in biomedical named entity recognition}. In
  \bibinfo{booktitle}{\emph{Machine Learning for Healthcare Conference}}.
  \bibinfo{pages}{383--402}.
\newblock


\bibitem[\protect\citeauthoryear{Saito, Watanabe, Ushiku, and Harada}{Saito
  et~al\mbox{.}}{2017}]%
        {mcd}
\bibfield{author}{\bibinfo{person}{Kuniaki Saito}, \bibinfo{person}{Kohei
  Watanabe}, \bibinfo{person}{Yoshitaka Ushiku}, {and} \bibinfo{person}{Tatsuya
  Harada}.} \bibinfo{year}{2017}\natexlab{}.
\newblock \showarticletitle{Maximum Classifier Discrepancy for Unsupervised
  Domain Adaptation}.
\newblock \bibinfo{journal}{\emph{CoRR}}  \bibinfo{volume}{abs/1712.02560}
  (\bibinfo{year}{2017}).
\newblock
\showeprint[arxiv]{1712.02560}
\urldef\tempurl%
\url{http://arxiv.org/abs/1712.02560}
\showURL{%
\tempurl}


\bibitem[\protect\citeauthoryear{Schwaighofer, Tresp, and Yu}{Schwaighofer
  et~al\mbox{.}}{2004}]%
        {schwaighofer2004learning}
\bibfield{author}{\bibinfo{person}{Anton Schwaighofer}, \bibinfo{person}{Volker
  Tresp}, {and} \bibinfo{person}{Kai Yu}.} \bibinfo{year}{2004}\natexlab{}.
\newblock \showarticletitle{Learning Gaussian process kernels via hierarchical
  Bayes}.
\newblock \bibinfo{journal}{\emph{Advances in neural information processing
  systems}}  \bibinfo{volume}{17} (\bibinfo{year}{2004}),
  \bibinfo{pages}{1209--1216}.
\newblock


\bibitem[\protect\citeauthoryear{Tan, Sun, Kong, Zhang, Yang, and Liu}{Tan
  et~al\mbox{.}}{2018}]%
        {deep_transfer_learning_survey}
\bibfield{author}{\bibinfo{person}{Chuanqi Tan}, \bibinfo{person}{Fuchun Sun},
  \bibinfo{person}{Tao Kong}, \bibinfo{person}{Wenchang Zhang},
  \bibinfo{person}{Chao Yang}, {and} \bibinfo{person}{Chunfang Liu}.}
  \bibinfo{year}{2018}\natexlab{}.
\newblock \showarticletitle{A Survey on Deep Transfer Learning}.
\newblock \bibinfo{journal}{\emph{CoRR}}  \bibinfo{volume}{abs/1808.01974}
  (\bibinfo{year}{2018}).
\newblock
\showeprint[arxiv]{1808.01974}
\urldef\tempurl%
\url{http://arxiv.org/abs/1808.01974}
\showURL{%
\tempurl}


\bibitem[\protect\citeauthoryear{Tang and Jia}{Tang and Jia}{2019}]%
        {dada}
\bibfield{author}{\bibinfo{person}{Hui Tang} {and} \bibinfo{person}{Kui Jia}.}
  \bibinfo{year}{2019}\natexlab{}.
\newblock \bibinfo{title}{Discriminative Adversarial Domain Adaptation}.
\newblock
\newblock
\showeprint[arxiv]{1911.12036}~[cs.CV]


\bibitem[\protect\citeauthoryear{Wu and Dietterich}{Wu and Dietterich}{2004}]%
        {wu2004improving}
\bibfield{author}{\bibinfo{person}{Pengcheng Wu} {and}
  \bibinfo{person}{Thomas~G Dietterich}.} \bibinfo{year}{2004}\natexlab{}.
\newblock \showarticletitle{Improving SVM accuracy by training on auxiliary
  data sources}. In \bibinfo{booktitle}{\emph{Proceedings of the twenty-first
  international conference on Machine learning}}. \bibinfo{pages}{110}.
\newblock


\bibitem[\protect\citeauthoryear{Yosinski, Clune, Bengio, and Lipson}{Yosinski
  et~al\mbox{.}}{2014}]%
        {yosinski2014transferable}
\bibfield{author}{\bibinfo{person}{Jason Yosinski}, \bibinfo{person}{Jeff
  Clune}, \bibinfo{person}{Yoshua Bengio}, {and} \bibinfo{person}{Hod Lipson}.}
  \bibinfo{year}{2014}\natexlab{}.
\newblock \showarticletitle{How transferable are features in deep neural
  networks?}. In \bibinfo{booktitle}{\emph{Advances in neural information
  processing systems}}. \bibinfo{pages}{3320--3328}.
\newblock


\end{thebibliography}

\end{document}